\documentclass{article}
\usepackage{arxiv}

\usepackage{times}  % DO NOT CHANGE THIS
\usepackage{helvet} % DO NOT CHANGE THIS
\usepackage{courier}  % DO NOT CHANGE THIS
\usepackage[hyphens]{url}  % DO NOT CHANGE THIS
\usepackage{graphicx} % DO NOT CHANGE THIS
\usepackage{graphicx}  % DO NOT CHANGE THIS
\usepackage{booktabs}       % professional-quality tables
\usepackage{amsfonts}% blackboard math symbols
\usepackage{nicefrac}% compact symbols for 1/2, etc.
\usepackage{microtype}% microtypography

\usepackage{amsmath}
\usepackage{array}
\usepackage{bm}
\usepackage{multirow}
\usepackage{algorithm}
\usepackage{algorithmic}
\usepackage{color}
\usepackage[position=top]{subfig}

\usepackage{indentfirst,amsmath,amssymb, comment, thm-restate, xpatch}

\newcommand{\eg} {\emph{e.g. }}

\newcommand{\w} {{ \mathbf{w} }}
\newcommand{\x} {{ \mathbf{x} }}

\newcommand*{\tran}{{^{\mkern-1.5mu\mathsf{T}}}}

\DeclareMathOperator{\relu}{ReLU}
\makeatother
\xpatchcmd{\thmt@restatable}% Edit \thmt@restatable
{\csname #2\@xa\endcsname\ifx\@nx#1\@nx\else[{#1}]\fi}% Replace this code
{\ifthmt@thisistheone\csname #2\@xa\endcsname\ifx\@nx#1\@nx\else[{#1}]\fi\else\csname #2\@xa\endcsname\fi}% with this code
{}{} % execute code for success/failure instances
\makeatother

\title{How Does BN Increase Collapsed Neural Network Filters?}
\author{
  Sheng Zhou\thanks{equal contribution} \space \thanks{Work done as a research intern in SenseTime} \\
  SenseTime Research\\
  \texttt{Thomaszhouan@gmail.com} \\
   \And
  Xinjiang Wang $^{\ast}$  \\
  SenseTime Research\\
  \texttt{wangxinjiang@sensetime.com} \\
   \And
  Ping Luo  \\
  The University of Hong Kong\\
  \texttt{pluo.lhi@gmail.com} \\
   \And
  Litong Feng \\
  SenseTime Research\\
  \texttt{fenglitong@sensetime.com} \\
   \And
  Wenjie Li $^{\dagger}$  \\
  SenseTime Research\\
  \texttt{williamleewj@gmail.com} \\
   \And
  Wei Zhang\\
  SenseTime Research\\
  \texttt{wayne.zhang@sensetime.com} \\
}
 
\begin{document}

\maketitle

\begin{abstract}
Improving sparsity of deep neural networks (DNNs) is essential for network compression and has drawn much attention. In this work, we disclose a harmful sparsifying process called filter collapse, which is common in DNNs with batch normalization (BN) and rectified linear activation functions (\eg ReLU, Leaky ReLU). It occurs even without explicit sparsity-inducing regularizations such as $L_1$. This phenomenon is caused by the normalization effect of BN, which induces a non-trainable region in the parameter space and reduces the network capacity as a result. This phenomenon becomes more prominent when the network is trained with large learning rates (LR) or adaptive LR schedulers, and when the network is finetuned.
We analytically prove that the parameters of BN tend to become sparser during SGD updates with high gradient noise and that the sparsifying probability is proportional to the square of learning rate and inversely proportional to the square of the scale parameter of BN.
To prevent the undesirable collapsed filters, we propose a simple yet effective approach named post-shifted BN (psBN), which has the same representation ability as BN while being able to automatically make BN parameters trainable again as they saturate during training.
With psBN, we can recover collapsed filters and increase the model performance in various tasks such as classification on CIFAR-10 and object detection on MS-COCO2017.
\end{abstract}
\section{Introduction}
\label{sec1}
Neural network sparsification as a method to accelerate its inference speed has
drawn much attention in the past years \cite{han_deep_2015,han_dsd:_2016,he_channel_2017,mao_exploring_2017,ide_improvement_2017,lee_compend:_2018}.
Commonly adopted methods for pruning either enforce an additional explicit
$L_1$ \cite{gordon_morphnet:_2018} or $L_0$ \cite{louizos_learning_2017}
regularization during training.% or manually prune parameters of low importance
%afterwards\cite{molchanov_pruning_2016}. 
%The sparsification imposed by the above methods are designed to be target-aware by comparing the contributions
%to the loss function from different parameters and pruning the smallest-contributing
%ones. Using these methods, network can be compressed by a great margin
%without hurting the test accuracy. 

In this study, we find that training a standard deep network such as ResNet \cite{he2016deep}  or VGG-BN \cite{simonyan2014very} can also generate structured network parameter sparsity without the above sparsity-imposing regularizers. The weight parameters of certain convolution kernels and the following BN parameters all become negligible (absolute value $<10^{-7}$) after training. We also find that one requisite for such phenomenon is the usage of batch normalization (BN)\cite{ioffe_batch_2015} before non-linear activation such as ReLU \cite{glorot_deep_2010,hahnloser_piecewise_1998} and Leaky ReLU \cite{maas2013rectifier}. 

As a successful normalization technique,
BN helps in different aspects in both training \cite{ioffe_batch_2015,santurkar2018does} and generalization \cite{luo2018towards}. However, its behavior in inducing a sparsified network has not been recorded in the literature. 

\begin{figure}
  \centering
    \subfloat[\label{fig:vgg_distribution}]{%
       \includegraphics[width=0.5\columnwidth]{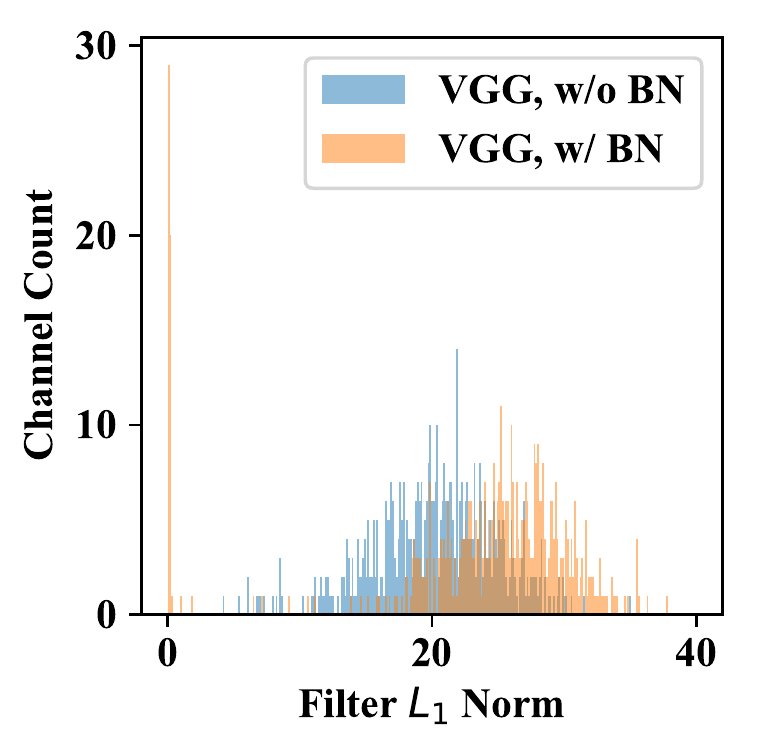}
     }
     \subfloat[\label{fig:param_trace}]{%
       \includegraphics[width=0.5\columnwidth]{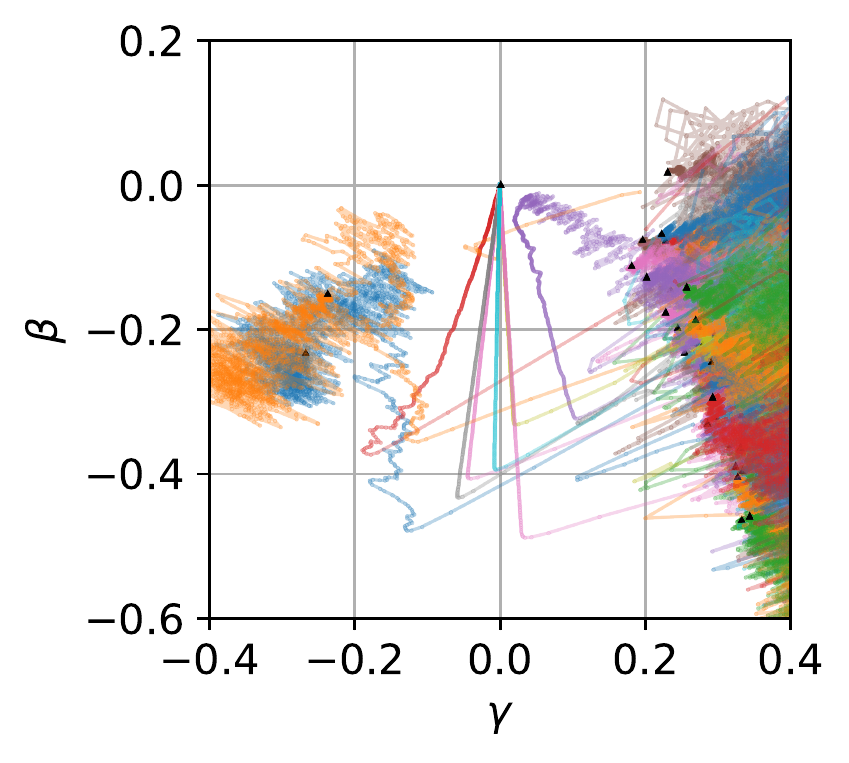}
     }
   \caption{(a) The $L_1$ norm distribution of convolutional filters of one layer for VGG11 with and without BN. %Both networks are trained on CIFAR-10 using SGD and cosine annealing learning rate with one restart \cite{loshchilov2016sgdr}.
   (b) The trace of BN scale ($\gamma$) and bias ($\beta$) parameters of one layer in ResNet-20 during the second round training on CIFAR-10. (Collapsed filters run from the outside to the origin point)}
  \label{fig:dead_cone}
\end{figure}

It is found in this work that the reason for such phenomenon lies in the combination of BN and rectified
nonlinear activation functions like ReLU. %, which is a standard setting in most current networks. 
When the network with BN is retrained on the same dataset using the same strategy, a much sparser network in the filter level is generated compared to the network without BN (Fig. \ref{fig:vgg_distribution}). 

This is because the normalization in BN implies a steady frozen zone in its parameter space of scale $\gamma$ and bias $\beta$. Once in the frozen space, the parameters will only be squashed to 0 by weight decay. This process is illustrated by the trace of BN parameters in Figure \ref{fig:param_trace}. We show such sparsification also exists in other network normalization methods as well, such as instance  normalization \cite{ulyanov2016instance}.  

\setlength\intextsep{0pt}

Some other contributing factors to the network sparsification are also discussed here. It is observed that a relatively large learning rate (LR), weight decay (WD) also contribute to the filter sparsity level. More importantly, iterative LR annealing schedules is also found to increase the sparsity in the network. This implies the wide-spread existence of collapsed filters in the network generated by iterative ``warmup'' \cite{loshchilov2016sgdr} schedules and common transfer learning tasks by network finetuning such as object detection \cite{lin2017focal}. 

With the finding of increased sparsity due to BN + ReLU, a natural question is whether the sparsity is favorable for the network.  
Several recent works treated the sparsity as a windfall in network training since it automatically generates a pruned network \cite{yaguchi_adam_2018,mehta_implicit_2018}.
However, in this work, we show that the accuracy of the sparsified
network only matches that using uniform channel pruning, suggesting detrimental effects of the collapsed filters.

Moreover, we also analytically model the transition probability of BN parameters to the non-trainable
regions for the first time. The BN parameters would always tend to
become sparser when the noise level of the gradient is large, which is common in training with large learning rates or small batches. From our statistical model, we also conclude that the sparsifying probability is proportional to the square of learning rate and inversely proportional to the square of the scale parameter of BN. These findings have been testified with experiments on training deep networks using natural datasets and help explain the increasing sparsity as a network is trained with iterative annealing of learning rate.

To escape the non-trainable region caused
by normalization and ReLU, we propose a simple yet effective method that only shifts the original BN by a small positive constant number. By utilizing this technique, the aforementioned
non-trainable region is proved non-stable, and collapsed filters would eventually be reactivated again. The representation
of the new normalization layer is mathematically equivalent to the original
one, and is able to increase the performance due to its removal of parameter
sparsity of BN. With this technique, a network is found to be able to defend common settings that would trigger collapsed filters. 
Apart from classification tasks, we also testify its usefulness in tasks with network parameters transfer by finetuning, such as
object detection on MS-COCO2017. A steady increase of mAP on a standard RetinaNet 
is also observed. 

This paper contributes in the following aspects: 
\begin{itemize}
\item We find that
the common combination of BN and ReLU induces collapsed filters in a network
and that the normalization operation in BN is the source of the stable non-tranable region. 
\item The sparsification due to BN + ReLU is found to actually harm both
training, generalization and transferability, which is undesirable for tasks aiming at higher performance. 
\item A mathematical model is provided
to explain the sparsification and predicts that the network will keep
sparsing during training with strong noises.
\item We propose a post-shifted BN (psBN) as an easy approach that not only avoids the network non-trainable region but still keeps sparse representations of ReLU;
\end{itemize}

\section{Related Work}

\label{sec2}
\subsection{Network pruning}

%Neuron networks are generally overparameterized and thus has room for
%pruning without hurting its performance. In practice,
%most methods adopt either explicit regularizations during training
%\cite{gordon_morphnet:_2018,louizos_learning_2017} or post-training
%pruning techniques by ranking the weight importance\cite{molchanov_pruning_2016}.

The sparsified representation of ReLU has inspired direct pruning
strategies on the feature map\cite{liang2018dynamic,liu_efficient_2018,dong2017activation} in the literature.% or sparse convolutions
%algorithms as an acceleration method\cite{liu_efficient_2018}. 
%Some other studies treated the magnitude output of ReLU as the importance
%metric for convolutional filter pruning\cite{li2016pruning,dong2017activation}.
%BN+ReLU is found in the current study to increase sparsification
%in parameter space as well. 
 However, there is a major difference between the collapsed filters found in this study from precedent
pruning methods utilizing ReLU sparsification. Whether or not BN induced sparsity is favorable for the loss function is still unknown. 
%still we do not know whether the awareness of loss functions during sparsification.
%Previous pruning algorithms mainly tried to minimize the increase
%of the loss while constraining a sparsity on the network.
In the current work, it is found that the network parameters almost always tend to
drop to the non-trainable zone, especially at a large
LR or with great gradient uncertainty. Therefore, instead of facilitating sparsity in network parameters, we present an appeal for cautious treatment of the collapsed filters. %And also we design anti-sparsification to help parameters escape the
%non-trainable zone during training, which would stretch the training
%ability of all parameters and thus increases the performance. 

\subsection{Dying ReLU}

Dying ReLU refers to a situation where the input of ReLU is
smaller than 0 and becomes non-trainable. 
%Despite its heuristic simplicity,
%it is still unclear with respect to the process, conditions
%and the significance of a neuron being dead. 
Several recent studies have
investigated the dying probability with network depth\cite{lu2019dying}.
The ReLU-related network sparsity in neural networks has also
been noticed by two contemporary works\cite{mehta_implicit_2018,yaguchi_adam_2018}, yet with no special attention to batch normalization.
The sparsification change with different hyper-parameters such
as LR, weight decay and training algorithms was experimentally analyzed
in \cite{mehta_implicit_2018}. \cite{yaguchi_adam_2018} also identified
the increased weight sparsity with Adam optimizer by relating it to
the larger convergence rate of Adam under only $L_2$ regularization when
weights are non-trainable by the target loss function. Both of these
two studies also tried to utilize it as a pruning methods. However,
in this study we doubt the efficacy of collapsed ReLU as a pruning method
by both comparing the performance of the sparsified model and random
pruning method and analytically illustrate the sparsing probability of
parameters with different distributions.

\subsection{Effects of BN in deep networks}
BN usually receives much applause due to its boosting performance in various tasks\cite{ioffe_batch_2015,szegedy2016rethinking}. It is generally thought to be able to allow a larger learning rate \cite{bjorck2018understanding}, 
 smooth the loss landscape \cite{santurkar2018does} and increase network generalization ability \cite{luo2018towards}.
On the other hand, \cite{yang2019mean} also noted a side effect of BN. It was derived that at random initialization, the inter-sample correlation due to BN could lead to gradient explosion, leading to non-trainable deep networks. Although the non-trainable filters due to BN+ReLU in this work may be reminiscent of gradient explosion due to BN, there are some key differences both empirically and theoretically. (1) The analysis in \cite{yang2019mean} mainly applies for \emph{random initialization}, when BN becomes ill-conditioned. However, the network with BN at random initialization is still trainable in this study. On the contrary, filter-level sparsity occurs when LR suddenly increases at convergence, where the gradients tend to be small instead of exploding. (2) The cause of filter sparsity in this study is analyzed from the stabilization effect of BN on feature distribution. Whereas the cause of gradient explosion in BN is the inter-sample correlation when batch size is small. 

\section{How does BN+ReLU make network parameters sparse?}

\subsection{Experimental setup}
\label{sec:cifar_exp}
We study the filter collapsing phenomenon on CIFAR-10 dataset \cite{CIFAR10} using three architectures with different depth and paradigm: ResNet-20, ResNet-56 \cite{he2016deep} and VGG16-BN \cite{VGG}.
All networks are trained using SGD with momentum $0.9$ and cosine annealing learning rate schedule \cite{loshchilov2016sgdr} with initial learning rate $0.1$ for 100 epochs. 
We choose batch size $128$ and weight decay $5\mathrm{e}{-4}$ throughout the experiments.
Data augmentation is performed as in \cite{he2016deep} unless specified. 
We consider structured sparsity in this study, and a convolutional filter is considered as pruned when its corresponding scaling parameter in the following BN layer is smaller than a certain threshold.
The threshold is chosen to be $1\mathrm{e}{-3}$ in our experiments since pruning at this level does not bring observable performance degradation (accuracy $\pm 0.03\%$) to the models.

\subsection{Sparsity in warm restart}

We consider training a network for \textit{multiple rounds}, where a round refers to one stage of cosine annealing of the learning rate.
For each network architecture, we use $5$ different random seeds and train the network for $9$ rounds, and calculate the sparsity (i.e. the percentage of reduced FLOPS \cite{gordon_morphnet:_2018}) and validation accuracy at the end of each round. 
It is worthy of noting that the relatively large total training budget mainly helps manifest the sparsification effect of BN+ReLU, and experiments using much shorter training steps (40 epochs per round and 5 rounds in total) have also been conducted (see later discussions). 
The result is shown in Fig. \ref{fig:retrain} (the orange curve).
It is clear from the figure that after each round of training, the sparsity of a network is increased.
Surprisingly, the removable channels also have a clear distinction with other channels in the parameter space.
Figure \ref{fig:vgg_distribution} shows the distribution of the $L_1$ norm of convolutional filters for a snapshot after round 3, the distribution has an isolated peak at $0$ which resembles the parameter distribution of a network trained with explicit sparsifying regularizers such as the LASSO \cite{LASSO}.

Accompanying the increasing sparsity is the deteriorating validation performance.
As proposed in \cite{loshchilov2016sgdr}, performing restart during optimization can improve the performance of the final network.
Presumably, training the model for more rounds should not harm its performance.
Our results show that this is not the case with the commonly used BN $+$ ReLU network, since more filters will collapse and the network capacity will be decreased.

\subsection{The effect of random labels}
\begin{figure}
\centering
    \subfloat[\label{fig:random_label_top1}]{%
       \includegraphics[width=0.5\linewidth]{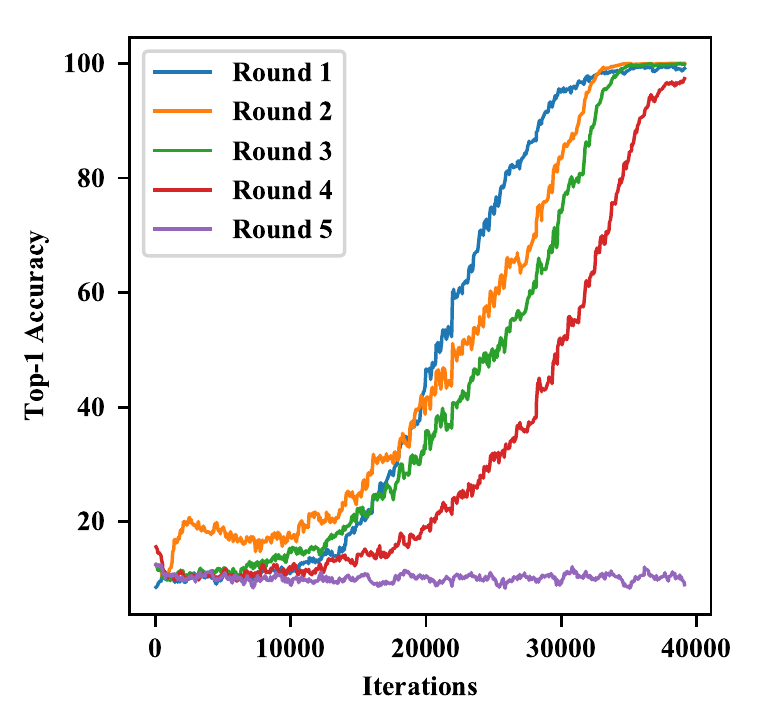}
     }
     \subfloat[\label{fig:random_label_flops}]{%
       \includegraphics[width=0.5\linewidth]{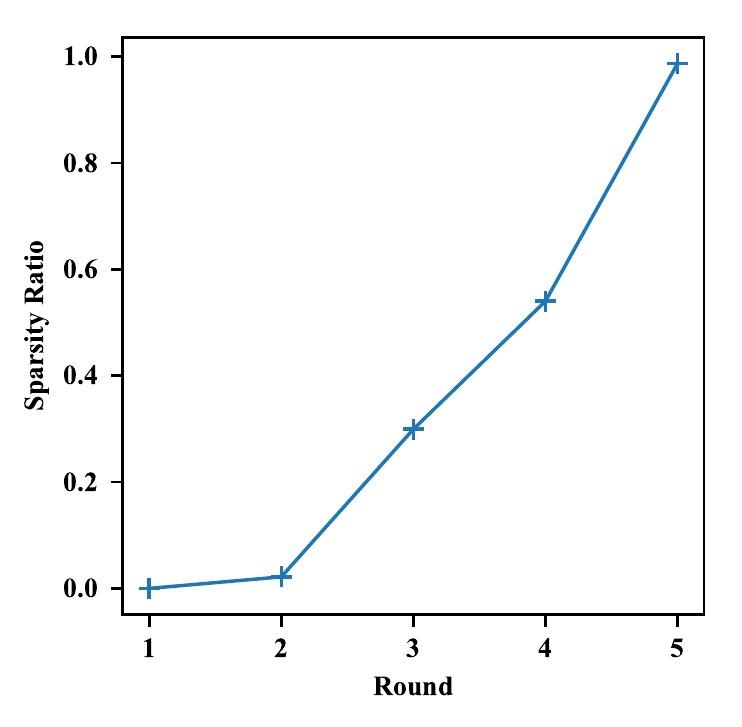}
     }
\caption{(a) Training curve of ResNet-20 on CIFAR-10 with random labels. The network fails to fit the training data during the fifth round due to filter collapse. (b) Sparsity ratio of ResNet-20 on CIFAR-10 with random labels. The sparsity increases with each round of restart.}
\end{figure}

In this subsection, we try to exclude the impact of a specific dataset by training a network with random labels. We take the CIFAR-10 dataset and randomly shuffle the image labels before the training starts.
Note that the correspondence between an image and its label is \textit{fixed} but kept \textit{random}. A network trained using such dataset will have minimal, if any, dependence on the latent structure of the data distribution, and a high performance network has to "remember" the dataset.
We train ResNet-20 on CIFAR-10 with random labels and the training loss of each round is shown in Fig. \ref{fig:random_label_top1}. In the first four rounds, the network fits the training data quite well, while its sparsity keeps increasing after each round of training (Fig. \ref{fig:random_label_flops}).
The presence of filter collapse on the random dataset shows that this phenomenon is not specific to a particular dataset.

\subsection{The effect of BN on filter collapse}

We study the effect of batch normalization on network sparsity caused by filter collapse.
The distribution of the filter $L_1$ norm for a specific layer in VGG11-BN at the end of the second round is shown in Fig. \ref{fig:vgg_distribution}.
The filter sparsity only becomes evident with the presence of BN. The distribution for VGG11-BN has a characteristic high peak at zero while the distribution of VGG11 does not have noticeable concentration near zero. Therefore collapsed filters should be related with BN. Moreover, the filter-level sparsity is not unique to BN but also observed in other normalizations, \eg instance normalization, suggesting a similar role played by the normalization during the process. 

\subsection{The effect of non-linear activation on filter collapse}

To study the effect of non-linear activation on filter collapse, we replace ReLU with Leaky ReLU with slope $0.01$ at the negative range. Leaky ReLU has a finite activation at the negative range. However, from the change in both network sparsity and accuracy change with different rounds of training (Fig. \ref{fig:retrain}), we observe that Leaky ReLU only has a marginal effect in avoiding collapsed filters, which also coincides with the conclusion in \cite{mehta_implicit_2018}. 
We have also tested the case where ReLU activations is removed after each BN layer of VGG11-BN, resulting in a deep linear network. The linear network is trained using the same strategy as before, and sparsity is barely observed ($<1\%$) at the end of the second round of training. Therefore, the phenomenon of collapsed filters is likely to be caused by a combined effect of BN and reticified activation functions. 

\subsection{Other factors contributing to filter collapse}
\begin{figure}[hbtp]
\centering
\includegraphics[width=\linewidth]{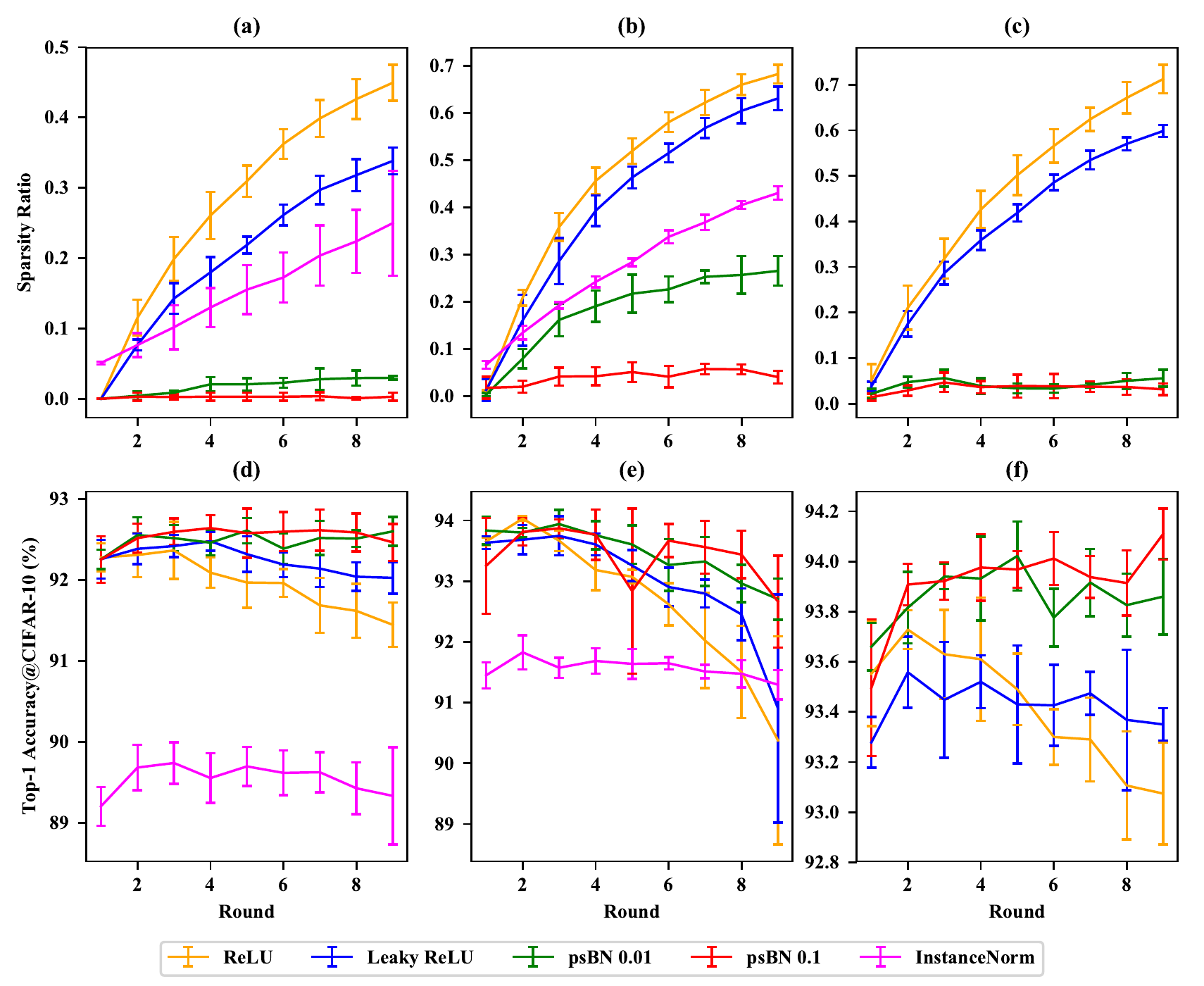}
\caption{Sparsity and accuracy comparison of different normalization-activation combinations. The horizontal axis is the number of training rounds.
Top row: sparsity ratio for (a) ResNet-20; (b) ResNet-56; (c) VGG16-BN.
Bottom row: top-1 accuracy on CIFAR-10 for (d) ResNet-20; (e) ResNet-56; (f) VGG16-BN.}
\label{fig:retrain}
\end{figure}

As shown in Fig. \ref{fig:retrain}, collapsed filters only appear starting from the \textit{second} round of training. 
Since the network architecture is invariant during learning rate restart, there must be other factors contributing to this phenomenon.

\paragraph{Filter selectivity?}
A reasonable hypothesis is that since the network has already converged at a good local minimum at the end of the first round of training, training the network for another round allows the network to select and prune unimportant filters by itself.
However, this assumption is disproved by our experiments.
We consider a trained network at the end of round 1 and randomly shuffle its parameters within the same parameter tensor.
If a network can self-prune based on a loss-aware selectivity metric (e.g. Fisher information, Hessian, etc), decorrelating network weights and loss should result in much fewer sparsity. Surprisingly, training the trained-and-shuffled network for another round still induces collapsed filters. This observation weakens the relation between network sparsification and loss-based selectivity.
In fact, the performance degradation observed even suggests that a trained model may \textit{not} be a good initialization at all.

\paragraph{Distribution of $\gamma$}
A common practice for initializing BN parameters is to set the scale to $1$ and the bias to $0$.
This is different from the distribution of $\gamma$s when the model is converged at the end of round 1, where most of the $\gamma$s are much smaller than $1$.
We hypothesize that the smaller initial value of $\gamma$s for subsequent rounds encourages the filters to collapse.
This is experimentally confirmed by setting $\gamma$s to smaller values at the \textit{first} round of training.
The sparsity of different networks for different values of initial $\gamma$s is shown in Fig. \ref{fig:GammaSparse}.
The sparsity of the network increases significantly when the initialization of $\gamma$s is decreased.
%Note that the threshold on $\gamma$s for measuring sparsity is fixed, but the comparison is still fair since the range of $\gamma$s for the non-collapsing filters are similar for different initializations.

\setlength\intextsep{0pt}
\begin{figure}
\centering
\includegraphics[width=0.7\linewidth]{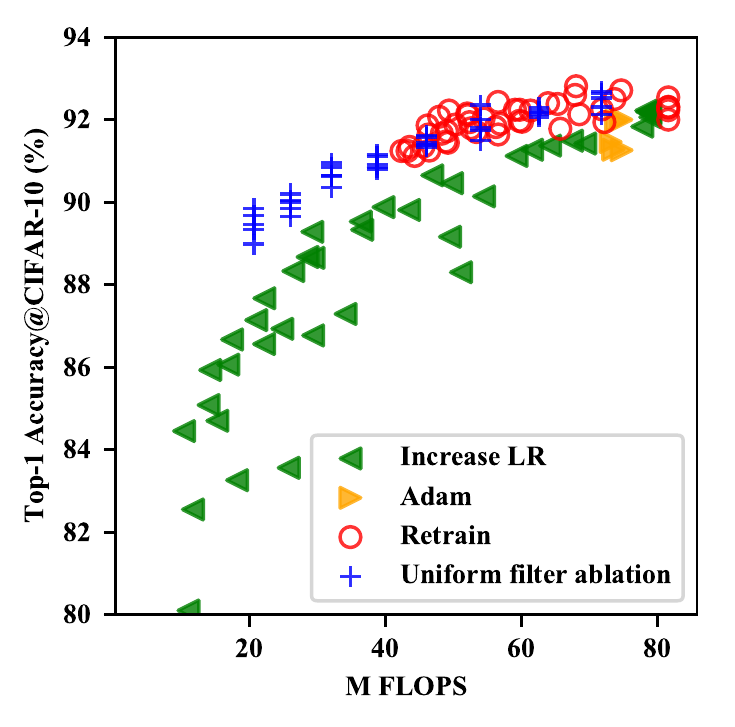}
\caption{Comparison of network pruning methods for ResNet-20 on CIFAR-10. We compare uniform filter ablation with methods that collapse filters using (1) larger LR; (2) adaptive LR method (Adam); (3) retrain for multiple rounds.}
\vspace{-5pt}
\label{fig:pruning_comparison}
\end{figure}

\subsection{Effectiveness of filter collapse for network pruning}
Since filter collapse introduces removable convolutional channels in the network, it is natural to consider exploiting this for network pruning. We compare the following sparsity-inducing methods: training the network for multiple rounds (\textbf{retrain}), using large learning rate (\textbf{Increase LR}), using Adam optimizer \cite{mehta_implicit_2018} and uniformly ablating all convolutional filters (\textbf{Uniform filter ablation}). All methods are applied to pruning ResNet-20 on CIFAR-10. (Detailed training settings can be seen in the supplementary material) 

The accuracy and computational cost (in FLOPS) for the pruned network using four different methods is shown in figure \ref{fig:pruning_comparison}. Interestingly, no method performs better than the naive \textbf{Uniform filter ablation} method. This comparison suggests that pruning the collapsed filters is not better than pruning random filters.

\section{Why does BN + ReLU make network parameters sparse?}
\label{sec3} 
In this section, we aim to find the conditions that drive a neuron to the non-activation range. %For a deep network, however, it is difficult to directly model the transition process. During experiments, 
We found that the majority of transition process of the BN parameters to the non-trainable range happens at the first several hundreds of steps when LR suddenly increases. The noise brought by the increased LR becomes so strong that it may even overwhelm the gradient propagated from the target loss function. Therefore, it would be helpful to find the relation between the transition probability of BN parameters to the non-trainable region and the noise level of the gradient. %However, there has been no metrics that measure the non-training probability for BN parameters yet. Instead, 
Here we use the activation probability of the neuron $\mathbb{P}(\gamma\hat{x} + \beta > 0)$ to also represent possibility for a set of parameters $\gamma$ and $\beta$ to be trainable. The BN parameters are updated at each SGD step and changes the neuron activation probability. 
%\subsection{A simplified illustration}
%As a simple analysis, we take the input of a layer as $\mathbf{x}$
%and the output after a single activation as $y$ and $y=\relu\left(\gamma\cdot\hat{x}+\beta\right)$,
%where $\gamma$ and $\beta$ are the scale and bias parameters of
%BN and 
%is the normalized version of $\tilde{x}=\mathbf{w}^{T}\mathbf{\cdot x}$
%with $\mu$ and $\sigma$ representing the average and standard deviation
%in one mini-batch. Since $\hat{x}$ is normalized, it is easy to see
%that $P\left(\hat{x}<-\delta\right)\leq\frac{1}{\delta^{2}}$ from
%Chebyshev inequality, where $\delta$ is positive and $P$ respresents
%the probability. Therefore, given $\beta/\left|\gamma\right|<-3$,
%the probability of activation would be smaller \textasciitilde{}0.1.
%Considering the loose bound of Chebyshev inequality, this neuron is
%highly likely to be inactive during most training steps.
\subsection{Notations and Assumptions}
In batch normalization that is applied in convolutional neural networks, the average statistics is calculated for a single filter without considering inter-channel correlation. Therefore it would ease the calculation by dropping the neuron index and focus on the behavior of a single neuron. The activation magnitude after ReLU is $y = \relu(\gamma\hat{x} + \beta)$. $\hat{x}$ is the normalized output after a linear kernel $\hat{x} = \frac{\w \tran \cdot \x -\mu_{\mathcal{{B}}}}{\sigma_{\mathcal{{B}}}}$, where $\mu_{\mathcal{{B}}}$ and $\sigma_{\mathcal{{B}}}$ are the batch average and standard deviation of the linear output $\w \tran \cdot \x$. Despite the form of a single neuron activation, it is able to represent behavior of all neurons in the same layer by assigning $\gamma$ and $\beta$ as variables drawn from specific unknown distributions $P(\gamma)$ and $P(\beta)$. Even though the distribution of input layer  $\x$ is unknown, the distribution of $\w \tran \cdot \x$ reaches Gaussian-alike especially as the number of input channels $N\rightarrow \infty$ due to the central limit theorem. This  applies for deeper layers in a neural network. %, as seen from \ref{•}the distribution of $\hat{x}$ in different layers. %At present we do not do such assumptions and 
 Let $\phi(\hat{x})$ represent the probability distribution function of $\hat{x}$ and let $\Phi(\hat{x})$ be its cumulative distribution.
  
\subsection{Neuron activation probability}

With the above treatments and assumptions, we are able to calculate the probability of a activating neuron. 
The probability that the layer is activated given parameters $\gamma$ and $\beta$ is defined as $\mathbb{P}(\gamma\hat{x} + \beta > 0)$
%\vspace{-5pt}
% \begin{equation}
%    \begin{aligned}
%     \mathbb{P}(\gamma\hat{x} + \beta > 0) = \mathbb{P}((\hat{x} > -\frac{\beta}{\gamma})) = 1 - \Phi(-\frac{\beta}{\gamma}) = \Phi(\frac{\beta}{\gamma})
%    \end{aligned}
% \end{equation}
Note that we assume that $\gamma > 0$ throughout the whole section without loss of generality since its behavior is symmetric for $\gamma<0$. 
After defining the variables and their probability density functions, we can examine the model  $y = \relu(\gamma\hat{x} + \beta)$ and its properties. Given $\eta$ and $L$ as the learning rate and the loss function, then the update rules are: 
\vspace{-5pt}
\begin{equation}
    \begin{aligned}
    \Delta\gamma &= -\eta\hat{x}\frac{\partial L}{\partial y} H(\gamma\cdot \hat{x}+\beta)\\
    \Delta \beta &= -\eta \frac{\partial L}{\partial y} H(\gamma\cdot \hat{x}+\beta)
    \end{aligned}
    \label{eq:update}
\end{equation}
\vspace{-2pt}
where $\frac{\partial L}{\partial y}$ is the gradient to the activation output and $H(\cdot)$ is the Heaviside step function. %With little information of the gradient $\frac{\partial L}{\partial y}$, it is difficult to reach any interesting conclusions. 
Here we focus our attention to the ``local'' behavior of the current layer, which cares only the gradient correlation between the neighboring two layers while treating the gradients from higher levels as independent.%, which echos the reverse Markov process. 
 Moreover, as stated above, the sparsification process is observed to happen when the noise overwhelms its true gradients. Therefore we treat  $\frac{\partial L}{\partial y}$ as an independent gradient with zero mean and variance $c^2$, where $c$ is a constant number. Then $\Delta \gamma$, $\Delta \beta$ are both random variables depending on $\frac{\partial L}{\partial y}$.  After a single step of update with stochastic gradient descent (SGD), BN parameters now become  $\gamma' = \gamma + \Delta\gamma, \beta' = \beta + \Delta \beta$. 

The probability that the layer is activated before the update is 
\vspace{-5pt}
 \begin{equation}
    \begin{aligned}
     \mathbb{P}(y>0) = \mathbb{P}((\hat{x} > -\frac{\beta}{\gamma})) = 1 - \Phi(-\frac{\beta}{\gamma})=\Phi(\frac{\beta}{\gamma})
    \end{aligned}
 \end{equation}
\vspace{-2pt}
 Therefore, the expectation for $\mathbb{P}(y > 0)$ is $\mathbb{E}_{\gamma, \beta}(\Phi(\frac{\beta}{\gamma}))$. After the update, the new expected probability for the layer activation is $\mathbb{E}_{\gamma\prime, \beta^\prime}(\Phi(\frac{\beta'}{\gamma'}))$. 
 %Since $\hat{x}$ is not changed during the update, we will omit it in both expectations for simplicity of notations. 
 We show the expected probability that the layer is activated becomes smaller during SGD updates by proposing the following theorem, which proves that the expected activation probability will tend to decrease at each step of learning, i.e. 
 \vspace{-5pt}
$$\mathbb{E}_{\gamma', \beta'}(\Phi(\frac{\beta'}{\gamma'}))< \mathbb{E}_{\gamma, \beta}(\Phi(\frac{\beta}{\gamma}))$$
\vspace{-2pt}
 
\begin{restatable}{theorem}{MultivariateTheorem}
\label{MultivariateTheorem}
Assume the update of BN parameters $\gamma$ and $\beta$ update according to Eq. (\ref{eq:update}), further assume that $\hat{x}\sim N(0,1)$, then  
\vspace{-5pt}
\begin{equation}
\label{eq:dead_zone_prob}
\mathbb{E}_{\gamma^\prime, \beta^\prime}(\Phi(\frac{\beta^\prime}{\gamma^\prime})) = \mathbb{E}_{\gamma, \beta}(\Phi(\frac{\beta}{\gamma})) + \frac{\eta^2c^2}{2}\int_{\mathbb{R}}\frac{1}{\gamma^2} P(\gamma) J(\gamma) d\gamma 
\end{equation}
where
\vspace{-5pt}
\begin{equation}
J(\gamma) =  \int_{\mathbb{R}} K(\frac{\beta}{\gamma}) P(\beta) d\beta
\end{equation}

and 
\begin{equation}
    \begin{aligned}
    K(\frac{\beta}{\gamma}) =  ((\frac{\beta}{\gamma})^4 - 2) \phi(\frac{\beta}{\gamma})^2 + (\frac{\beta}{\gamma}-(\frac{\beta}{\gamma})^3)\phi(\frac{\beta}{\gamma})\Phi(\frac{\beta}{\gamma})
    \end{aligned} 
\end{equation} 
\vspace{-2pt}
Further assume $\beta$ follows an even distribution, $J(\gamma)$ is then a function that is always negative and hence $\mathbb{E}_{\gamma', \beta'}(\Phi(\frac{\beta'}{\gamma'}))< \mathbb{E}_{\gamma, \beta}(\Phi(\frac{\beta}{\gamma}))$

\end{restatable}

We provide detailed proof in the supplementary. The above theorem provides a proof that neurons are prone to be less activated during each updating step at a high noise level of gradients, and it provides some key insights towards understanding the sparsification process. 
\paragraph{Learning rate and gradient noise} 
According the Eq. (\ref{eq:dead_zone_prob}), it is noticed that the decrease of neuron activation probability  is proportional to  $\eta^2c^2$, where $\eta$ is the learning rate and $c$ is the variance of the gradients in regards to the output. This relation reveals that the noise in gradients might drives BN parameters into the non-trainable region, and that large learning rate behaves similarly with to noise level.

\begin{figure}
\center
\subfloat[\label{fig:LRSparsity}]{
\includegraphics[width=0.5\linewidth]{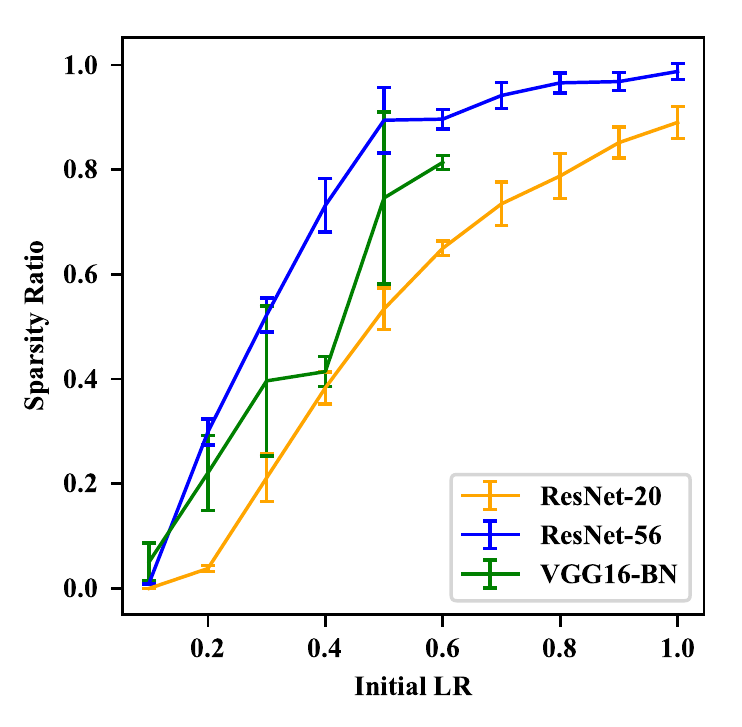}}
\subfloat[\label{fig:GammaSparse}]{
\includegraphics[width=0.5\linewidth]{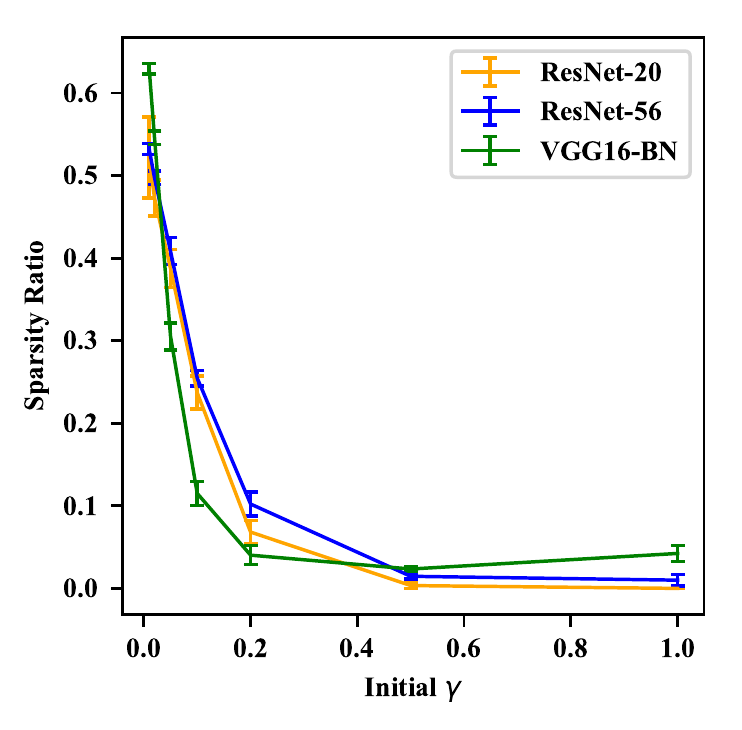}}
\caption{The sparsity level change with (a) LR and (b) $\gamma$ initialization}
\end{figure}

To study the relation between learning rate and the level of filter collapse, we train networks on CIFAR-10 with initial learning rate in range $[0.1, 1]$. The resulting sparsity is shown in figure \ref{fig:LRSparsity}. It can be seen that larger learning rate induces more collapsed filters.

\paragraph{Initialization of $\gamma$} 
For a general initialization technique of BN, $\beta$ is initialized to be $0$. In this scenario, it is easy to prove that $J(\gamma)=C$, where $C$ is a negative constant. Therefore, the decrease of neuron activation is proportional to $\frac{1}{\gamma^2}$. 
Fig. \ref{fig:GammaSparse} shows the change of network that the sparsity level $\gamma$ initialization value. 
Therefore, the conclusions from Theorem. \ref{MultivariateTheorem} are verified experimentally.% by the increased sparsity level with both learning rate and decreasing $\gamma$ initialization values. 

\setlength\intextsep{0pt}

\subsection{Weight decay}
The above analysis discusses the transition probability of BN parameters to the non-trainable region. %However, even though the activation probability $\mathbb{P}(y>0)$ approaches 0, BN parameters could also jump out of the non-trainable zone given long enough time. Therefore, 
It is also found that weight decay does more than an equivalent $L_2$ regularization that helps avoid overfitting. Weight decay is also able to collect all non-trainable parameters to zero. Moreover, under a constant weight decay factor $\lambda$. It would be easy to derive that the ratio $\frac{\beta}{\gamma}$ would maintain the original value throughout the decay. It indicates an almost intact non-activating status for these two parameters and  both parameters in BN finally approaches 0, representing a true sparsified filter. 

\section{Proposed method of avoiding sparsity}
In the above sections, BN + ReLU has been shown to induce collapsed filters under various conditions. Filters always tend to drop to the non-trainable region regardless of the training target when learning rate is large. Therefore, avoiding the non-trainable region is desirable in order to fully stretch the training ability of each parameter in the network. A direct solution would be adding a regularization term to the training loss. However, adding regularizations highly relies on prior knowledge of the distribution of network parameters. %An improper regularization might induce unexpected training constraints as well. 
In this study, we propose a more elegant method that adopt a post-shifted BN that can solve the collapsing filter problem more elegantly. 
\vspace{-3pt}
\begin{equation}
y = \text{BN}(\hat{x})+\alpha = \gamma\cdot \frac{\hat{x} -\mu_{\mathcal{{B}}}}{\sqrt{\sigma_{\mathcal{{B}}}^2 + \epsilon}} + \beta + \alpha \quad\quad (\alpha>0)\\
\end{equation}
where the definition of all parameters are the same as in the original batch normalization \cite{ioffe_batch_2015} except that a small positive constant $\alpha$ is added to prevent collapsed filters. Since $\alpha$ is a mere constant, using it to shift BN is mathematically equivalent to the original one. However, the BN parameters would behave differently as they fall into the non-trainable region at a step. 
To demonstrate the difference, we assume there exist a BN filter at time step $t$ satisfying $\frac{\beta^t+\alpha}{\vert\gamma^t\vert}=C^t\ll 0$, indicating a collapsed filter. %meaning that the filter is collapsed with slim probability to be activated.
After a single step of weight decay ($\Delta \gamma^t=-\eta\lambda\gamma^t$ and $\Delta \beta^t=-\eta\lambda\beta^t$, where $\lambda$ is weight decay strength), it can be easy to show  $C^{t+1}=C^t+\frac{\eta\lambda}{1-\eta\lambda}\frac{\alpha}{\vert\gamma\vert}>C^t$. Therefore, The collapsed filter would sooner be reactivated instead of decaying directly to $0$. On the other hand, the $\alpha$ should be small enough to avoid a strong prior on the bias of BN. It has been tested on different tasks that $\alpha=0.1$ works stably for different tasks and is thus mostly adopted in the following discussion. 

\subsection{Avoiding sparsity in CIFAR-10 training}
In this section, we show that using psBN can prevent filter collapse for networks trained with cyclic learning rate on CIFAR-10 dataset. As for the experimental setup in Fig. \ref{fig:retrain}, we replace the BN layers of a network with psBN using $\alpha=0.1$, and compare psBN against the BN + ReLU/Leaky ReLU baselines. As shown in the figure, psBN  inhibits filter collapse and performance degradation significantly especially in the later rounds.

\subsection{Avoiding sparsity in snapshot ensemble}
\label{subsec:snapshot}
As proposed in \cite{SnapshotEnsemble}, snapshots of a network at the end of each round of learning rate decay can be ensembled to boost the test time performance.
%We have also testified the performance of psBN when applied to the snapshot ensemble method\cite{SnapshotEnsemble}. 
Following the original setup, the total training budget is constrained to 200 epochs and is divided into 5 cyclic rounds. Then the models of the last three snapshots are ensembled to reach higher performance than the model directly trained in a single LR decay using the same training budget (200 epochs).  The baseline top-1 accuracy of ensembles for initial LR equaling 0.1 \& 0.2 are \textbf{94.27} \& \textbf{94.68} from the original paper. As seen from Table \ref{tab:cifar-ensemble}, the network sparsity also nearly monotonically increases for a standard snapshot training. Thus the representation power of the ensembled model is not fully utilized due to the filter collapse phenomenon. This problem is alleviated by psBN, which evidently outperforms the baseline of snapshot ensemble.

\begin{table*}
\caption{Snapshot Ensemble \cite{SnapshotEnsemble} training for ResNet-110 on CIFAR-10. Each cell is in the format \textit{psBN(baseline)}.}
\label{tab:cifar-ensemble} \centering %
\small

        \begin{tabular}{cccccccc}
\toprule 
InitLR & Metric & \#1 & \#2 & \#3 & \#4 & \#5 & Ensemble  \tabularnewline 
\hline 
0.2 & Val acc (\%) & 92.38(92.23) & 93.51(93.51) & 93.92(93.94) & 94.00(94.01) & 94.29(94.10) & \textbf{94.86}(94.63) \\ 
0.2 & Sparsity (\%) & 0.4(0.4) & 1.0(8.1) & 2.8(16.6) & 4.5(24.5) & 5.3(31.6) & N/A \\ 
\hline 
0.1 & Val acc(\%) & 88.86(91.26) & 92.81(92.93) & 93.49 (93.18) & 94.05 (93.67) & 94.43(93.75) & \textbf{94.75}(94.12) \\ 
0.1 & Sparsity (\%) & 0.2 (0.0) & 0.2(0.0) & 0.8(1.2) & 0.6(2.8) & 0.8(3.8) & N/A \\ 
\bottomrule
\end{tabular} 
\end{table*}

\subsection{Avoiding sparsity in COCO-2017 Object Detection}
From previous discussions, iterative LR annealing would result in more sparsity to the network and avoiding the undesirable sparsity brings a performance increase. Extending the observation and our theoretical proof to a broader span, one would find that a sudden learning rate increase also exist in various transfer learning tasks including object detection, where neural networks which are often pretrained on ImageNet are finetuned again. Therefore, a certain degree of sparsity would also occur during the finetuning process. 

We experiment on the RetinaNet\cite{lin2017focal} baseline using a ResNet-50 \cite{he2016deep} backbone. When examining the distribution of BN weights of the model trained with standard BN+ReLU, we also found sparsified weights in deep layers, suggesting possibly collapsed ReLU.  

Table \ref{tab:coco} shows the comparison between our proposed shifted BN and both the reported and our implemented baseline. The shifted BN shows a steady increase over the performance of RetinaNet across different image scales.

\begin{table}
\caption{Post-shifted BN (psBN) performance on MS-COCO2017 dataset. The baseline performance of RetinaNet\cite{lin2017focal} reported in the original paper is also listed in the bracket following the AP value of our implementation.}
\label{tab:coco} \centering %
\begin{tabular}{lll}
\toprule 
 Image scale  & AP$^{bbox}$ (baseline) & AP$^{bbox}$ (psBN) \tabularnewline 
 500  & 32.7(32.5)& 33.4 \tabularnewline
 800  & 35.8(35.7) & 36.4 \tabularnewline
\bottomrule
\end{tabular}
\end{table}

\section{Conclusion}
In this study, we report a side effect of the common combination between BN and ReLU. The normalization layer in BN induces a stable non-trainable region in BN parameters due to the increased collapsing ReLU. Apart from the larger learning rate (LR) and adaptive LR optimization schedules that also contribute to more sparse parameters, we also found that this phenomenon is more manifest with multiple rounds of annealing LR, and that the induced sparsity also exist in other normalizations such as instance normalization. The phenomenon is attributed to the stable non-trainable zone for parameters induced by the normalization in BN. Instead of being potential to be a pruning method, the sparsity in the network is found to be harmful to the network performance. We also analytically prove that the network parameters tend to become more sparse at each SGD update. A post-shifted BN (psBN) is also proposed to help network escape the undesirable sparsity. It is advantageous in having the same representation ability with BN and being able to make parameters retrainable again. Its efficacy have been testified on both CIFAR-10 and MS-COCO2017 datasets with stable increase in the performance.  

\clearpage

\bibliographystyle{unsrt} 
\bibliography{main}

\end{document}

\maketitle
\onecolumn

In this supplementary material, we provide proof for the main theorem stated in our paper, as well as detailed settings for the experiments.

\section{Proof of the main theorem}

\begin{lemma}
\label{scalelemma}
Let $X$ be a random variables with probability density function $f_X$, $a$ be a constant, then the density function for random variable $Z = aX$ is $f_{Z}(z) = \frac{1}{|a|} f_X(\frac{z}{a})$
\end{lemma}
\emph{\textbf{Proof:}}
\begin{equation}
    \mathbb{P}(Z\leq z) = 
    \begin{cases}
        \mathbb{P}(X\leq \frac{z}{a}), \text{if $a > 0$} \\
        \mathbb{P}(X\geq \frac{z}{a}), \text{if $a < 0$}
    \end{cases}
\end{equation}

Hence take derivative with respect to z we have, 

\begin{equation}
    f_{Z}(z) = 
    \begin{cases}
        \frac{1}{a}f_X(\frac{z}{a}), \text{if $a > 0$} \\
        -\frac{1}{a}f_X(\frac{z}{a}), \text{if $a < 0$}
    \end{cases}
    = \frac{1}{|a|}f_X(\frac{z}{a})
\end{equation}

% \MultivariateTheorem*
% Replace here!!!
\begin{restatable}{theorem}{MultivariateTheorem}
\label{MultivariateTheorem}
Assume the update of BN parameters $\gamma$ and $\beta$ update according to Eq. 2 of the main text, further assume that $\hat{x}\sim N(0,1)$, then  
\begin{equation}
\label{eq:dead_zone_prob}
\mathbb{E}_{\gamma^\prime, \beta^\prime}(\Phi(\frac{\beta^\prime}{\gamma^\prime})) = \mathbb{E}_{\gamma, \beta}(\Phi(\frac{\beta}{\gamma})) + \frac{\eta^2c^2}{2}\int_{\mathbb{R}}\frac{1}{\gamma^2} P(\gamma) J(\gamma) d\gamma 
\end{equation}
where

\begin{equation}
J(\gamma) =  \int_{\mathbb{R}} K(\frac{\beta}{\gamma})P(\beta) d\beta
\end{equation}

and 
\begin{equation}
    \begin{aligned}
    K(\frac{\beta}{\gamma}) =  ((\frac{\beta}{\gamma})^4 - 2) \phi(\frac{\beta}{\gamma})^2 + (\frac{\beta}{\gamma}-(\frac{\beta}{\gamma})^3)\phi(\frac{\beta}{\gamma})\Phi(\frac{\beta}{\gamma})
    \end{aligned} 
\end{equation} 

Further assume $\beta$ follows an even distribution, $J(\gamma)$ is then a function that is always negative and hence $\mathbb{E}_{\gamma', \beta'}(\Phi(\frac{\beta'}{\gamma'}))< \mathbb{E}_{\gamma, \beta}(\Phi(\frac{\beta}{\gamma}))$

\end{restatable}

\emph{\textbf{Proof:}} Let the multivariate function $\psi(\gamma,\beta)=\Phi(\frac{\beta}{\gamma})$. Let $D(\psi)$ be its Jacobian matrix and $H(\psi)$ be the Hessian, then

\begin{equation}
D(\psi) =
  \begin{bmatrix}
    \phi(\frac{\beta}{\gamma})(-\frac{\beta}{\gamma^2}) &  \phi(\frac{\beta}{\gamma})(\frac{1}{\gamma})
  \end{bmatrix},
H(\psi) =
  \begin{bmatrix}
    \phi^\prime(\frac{\beta}{\gamma})(\frac{\beta}{\gamma^2})^2+\phi(\frac{\beta}{\gamma})(\frac{2\beta}{\gamma^3}) & \phi^\prime(\frac{\beta}{\gamma})(-\frac{\beta}{\gamma^3})+\phi(\frac{\beta}{\gamma})(-\frac{1}{\gamma^2})  \\
    \phi^\prime(\frac{\beta}{\gamma})(-\frac{\beta}{\gamma^3})+\phi(\frac{\beta}{\gamma})(-\frac{1}{\gamma^2}) & \phi^\prime(\frac{\beta}{\gamma})(\frac{1}{\gamma^2})\\
  \end{bmatrix} 
\end{equation}

Given $\hat{x}$, the second order Taylor Expansion for $\psi(\gamma,\beta)$ is:

\begin{equation}
    \begin{aligned}
    \psi(\gamma^\prime,\beta^\prime)
    & \approx \psi(\gamma,\beta)+ D(\psi) [\Delta\gamma, \Delta\beta]^T + \frac{1}{2}[\Delta\gamma, \Delta\beta]H(\psi)[\Delta\gamma, \Delta\beta]^T \\
    & = \Phi(\frac{\beta}{\gamma}) + \phi(\frac{\beta}{\gamma}) (-\frac{\beta}{\gamma^2}\Delta\gamma + \frac{1}{\gamma}\Delta\beta) + \{\frac{1}{2}\Delta\gamma^2[\phi^\prime(\frac{\beta}{\gamma})(\frac{\beta}{\gamma^2})^2+\phi(\frac{\beta}{\gamma})(\frac{2\beta}{\gamma^3})] \\ & +  \Delta\beta\Delta\gamma[\phi^\prime(\frac{\beta}{\gamma})(-\frac{\beta}{\gamma^3})+\phi(\frac{\beta}{\gamma})(-\frac{1}{\gamma^2})]  + \frac{1}{2}\Delta\beta^2   \phi^\prime(\frac{\beta}{\gamma})\frac{1}{\gamma^2}\} \\
    & = \Phi(\frac{\beta}{\gamma}) + \phi(\frac{\beta}{\gamma})(-\frac{\beta}{\gamma^2}\Delta\gamma + \frac{1}{\gamma}\Delta\beta + \frac{\beta}{\gamma^3}\Delta\gamma^2 - \frac{1}{\gamma^2}\Delta\beta\Delta\gamma) \\ & + \phi^\prime(\frac{\beta}{\gamma})(\frac{\beta^2}{2\gamma^4}\Delta\gamma^2- \frac{\beta}{\gamma^3}\Delta\beta\Delta\gamma+  \frac{1}{2\gamma^2}\Delta\beta^2) \\
    & = \Phi(\frac{\beta}{\gamma}) + \phi(\frac{\beta}{\gamma})(-\frac{\hat{x}\beta}{\gamma^2}\Delta\beta + \frac{1}{\gamma}\Delta\beta + \frac{\hat{x}^2\beta}{\gamma^3}\Delta\beta^2 - \frac{\hat{x}}{\gamma^2}\Delta\beta^2) \\ & + \phi^\prime(\frac{\beta}{\gamma})(\frac{\hat{x}^2\beta^2}{2\gamma^4}\Delta\beta^2- \frac{\hat{x}\beta}{\gamma^3}\Delta\beta^2+  \frac{1}{2\gamma^2}\Delta\beta^2) \\
    & = \Phi(\frac{\beta}{\gamma}) +  \phi(\frac{\beta}{\gamma})(-\frac{\hat{x}\beta}{\gamma^2} + \frac{1}{\gamma})\Delta\beta +  [\phi(\frac{\beta}{\gamma})(\frac{\hat{x}^2\beta}{\gamma^3} - \frac{\hat{x}}{\gamma^2}) + \phi^\prime(\frac{\beta}{\gamma})(\frac{\hat{x}^2\beta^2}{2\gamma^4}- \frac{\hat{x}\beta}{\gamma^3}+  \frac{1}{2\gamma^2})]\Delta\beta^2 \\
    & = \mathbb{S + T}
    \end{aligned}
\end{equation}

where $\mathbb{S}$ contains the items with $\hat{x}$ and $\mathbb{T}$ contains those without,

\begin{equation}
    \begin{aligned}
     \mathbb{S} & =  \phi(\frac{\beta}{\gamma})(-\frac{\hat{x}\beta}{\gamma^2} )\Delta\beta +  [\phi(\frac{\beta}{\gamma})(\frac{\hat{x}^2\beta}{\gamma^3} - \frac{\hat{x}}{\gamma^2}) + \phi^\prime(\frac{\beta}{\gamma})(\frac{\hat{x}^2\beta^2}{2\gamma^4}- \frac{\hat{x}\beta}{\gamma^3} )]\Delta\beta^2 \\
     \mathbb{T} & = \Phi(\frac{\beta}{\gamma}) +  \phi(\frac{\beta}{\gamma}) \frac{1}{\gamma}\Delta\beta + \phi^\prime(\frac{\beta}{\gamma}) \frac{1}{2\gamma^2}\Delta\beta^2 \\
    \end{aligned}
\end{equation}

Then the expectation is 

\begin{equation}
    \begin{aligned}
     \mathbb{E}_{\gamma^\prime, {\beta}^\prime}(\Phi(\frac{{\beta}^\prime}{\gamma^\prime})) &=\iiiint_{\mathbb{R}^4}\Phi(\frac{\beta^\prime}{\gamma^\prime})P(\Delta\beta|\gamma,\beta, \hat{x})P(\gamma,\beta)\phi(\hat{x}) d\hat{x} d\Delta\beta d\gamma d\beta \\ 
     &= \iiiint_{\mathbb{R}^4}(\mathbb{S+T})P(\Delta\beta|\gamma,\beta, \hat{x})P(\gamma,\beta)\phi(\hat{x}) d\hat{x} d\Delta\beta d\gamma d\beta
    \end{aligned}
\end{equation}

By Lemma \ref{scalelemma}, we can calculate the integration of $\mathbb{T}$:

\begin{equation}
    \begin{aligned}
     & \iiiint_{\mathbb{R}^4}\mathbb{T}P(\Delta\beta|\gamma,\beta, \hat{x})P(\gamma,\beta)\phi(\hat{x}) d\hat{x} d\Delta\beta d\gamma d\theta \\
     & = \iiint_{\mathbb{R}^3}\mathbb{T}P(\gamma,\beta)[\int_{-\infty}^{-\frac{\beta}{\gamma}} P(\Delta\beta|\gamma,\beta, \hat{x}) \phi(\hat{x})d\hat{x} + \int_{-\frac{\beta}{\gamma}}^{+\infty}P(\Delta\beta|\gamma,\beta, \hat{x}) \phi(\hat{x}) d\hat{x}] d\Delta\beta d\gamma d\beta\\
    \end{aligned}
\end{equation}

Note that given $\hat{X} = \hat{x}$, when $\hat{x} < -\theta, \Delta\gamma, \Delta\beta$ = 0, when $\hat{x} > -\theta$, $\Delta\gamma, \Delta\theta$ depend on the random variable $R$, Hence:
     
\begin{equation}
    \begin{aligned}
      &\int_{-\infty}^{-\frac{\beta}{\gamma}} P(\Delta\beta|\gamma,\beta, \hat{x}) \phi(\hat{x})d\hat{x}  = \int_{-\infty}^{--\frac{\beta}{\gamma}} \delta_{\Delta\beta} \phi(\hat{x})d\hat{x}
       = \delta_{\Delta\beta} \Phi(-\frac{\beta}{\gamma}) \\
      &\int_{-\frac{\beta}{\gamma}}^{+\infty}P(\Delta\beta|\gamma,\beta, \hat{x}) \phi(\hat{x}) d\hat{x} = \eta^{-1}f(-\frac{\Delta\beta}{\eta})\Phi(\frac{\beta}{\gamma}) d\hat{x}
     \end{aligned}
\end{equation}

Therefore the original integration of $\mathbb{T}$ now becomes

\begin{equation}
    \begin{aligned}
    & \iiiint_{\mathbb{R}^4}\mathbb{T}P(\Delta\beta|\gamma,\beta, \hat{x})P(\gamma,\beta)\phi(\hat{x}) d\hat{x} d\Delta\beta d\gamma d\beta \\
     & =  \iiint_{\mathbb{R}^3}\mathbb{T}P(\gamma,\beta)(\delta_{\Delta\beta}\Phi(-\frac{\beta}{\gamma}) + \eta^{-1}f(-\frac{\Delta\beta}{\eta})\Phi(\frac{\beta}{\gamma}))d\Delta\beta d\gamma d\beta\\
     & =  \iint_{\mathbb{R}^2}P(\gamma,\beta)\int_{\mathbb{R}}\mathbb{T}(\delta_{\Delta\beta}\Phi(-\frac{\beta}{\gamma}) + \eta^{-1}f(-\frac{\Delta\beta}{\eta})\Phi(\frac{\beta}{\gamma}))d\Delta\beta d\gamma d\beta\\
     & =  \mathbb{I} + \mathbb{II} + \mathbb{III} 
    \end{aligned}
\end{equation}

where we partition the integration into three parts

\begin{equation}
    \begin{aligned}
    \mathbb{I} & = \iint_{\mathbb{R}^2}P(\gamma,\beta)\int_{\mathbb{R}}\Phi(\frac{\beta}{\gamma})(\delta_{\Delta\beta}\Phi(-\frac{\beta}{\gamma})+\eta^{-1}f(-\frac{\Delta\beta}{\eta})\Phi(\frac{\beta}{\gamma}))d\Delta\beta d\gamma d\beta \\
     \mathbb{II} & = \iint_{\mathbb{R}^2}P(\gamma,\beta)\int_{\mathbb{R}}\phi(\frac{\beta}{\gamma}) \frac{1}{\gamma}\Delta\beta(\delta_{\Delta\beta}\Phi(-\frac{\beta}{\gamma})+\eta^{-1}f(-\frac{\Delta\beta}{\eta})\Phi(\frac{\beta}{\gamma}))d\Delta\beta d\gamma d\beta  \\
     \mathbb{III} & = \iint_{\mathbb{R}^2}P(\gamma,\beta)\int_{\mathbb{R}}\phi^\prime(\frac{\beta}{\gamma}) \frac{1}{2\gamma^2}\Delta\beta^2(\delta_{\Delta\beta}\Phi(-\frac{\beta}{\gamma})+\eta^{-1}f(-\frac{\Delta\beta}{\eta})\Phi(\frac{\beta}{\gamma}))d\Delta\beta d\gamma d\beta  \\
  \end{aligned}
\end{equation}

where $\delta_{x}$ is the Dirac delta function, 

\begin{equation}
\delta_{x} =\begin{cases}
      0, \text{if $x \neq 0$} \\
      +\infty, \text{if $x = 0$}
     \end{cases}
\end{equation}

The Dirac delta function satisfies the following properties which are used in the integration:
\begin{equation}
\begin{aligned}
\int_{-\infty}^{+\infty} \delta_{x} & = 1 \\
\int_{-\infty}^{+\infty} f(x)\delta_{x} & = f(0),  \forall f
     \end{aligned}
\end{equation}

Integrate on $\Delta \beta$:

\begin{equation}
    \begin{aligned}
\mathbb{I} & =  \iint_{\mathbb{R}^2}P(\gamma,\beta)\Phi(\frac{\beta}{\gamma})(\Phi(-\frac{\beta}{\gamma}) + \Phi(\frac{\beta}{\gamma}))d\Delta\beta d\gamma d\beta  \\
    & =  \iint_{\mathbb{R}^2}P(\gamma,\beta) \Phi(\frac{\beta}{\gamma}) d\gamma d\beta \\
 \mathbb{II} & = \iint_{\mathbb{R}^2}P(\gamma,\beta)\int_{\mathbb{R}}\phi(\frac{\beta}{\gamma}) \frac{1}{\gamma}\Delta\beta(\delta_{\Delta\beta}\Phi(-\frac{\beta}{\gamma})+\eta^{-1}f(-\frac{\Delta\beta}{\eta})\Phi(\frac{\beta}{\gamma}))d\Delta\beta d\gamma d\beta \\
  & = \iint_{\mathbb{R}^2} P(\gamma,\beta) \phi(\frac{\beta}{\gamma}) \frac{-\eta}{\gamma} \int_{\mathbb{R}} mf(m) dm d\gamma d\theta \\
  & = 0 \\
 \mathbb{III} & = \iint_{\mathbb{R}^2}P(\gamma,\beta)\int_{\mathbb{R}}\phi^\prime(\frac{\beta}{\gamma}) \frac{1}{2\gamma^2}\Delta\beta^2\eta^{-1}f(-\frac{\Delta\beta}{\eta})\Phi(\frac{\beta}{\gamma})d\Delta\beta d\gamma d\beta\\ 
 & =  \frac{\eta^2}{2} \iint_{\mathbb{R}^2}\Phi(\frac{\beta}{\gamma})P(\gamma,\beta)\phi^\prime(\frac{\beta}{\gamma})\int_{\mathbb{R}} m^2 f(m)d m \\
 & = \frac{\eta^2c^2}{2} \iint_{\mathbb{R}^2}P(\gamma,\beta) \Phi(\frac{\beta}{\gamma})\phi^\prime(\frac{\beta}{\gamma})\frac{1}{\gamma^2} d\gamma d\beta
  \end{aligned}
\end{equation}

Hence
\begin{equation}
    \begin{aligned}
  & \iiiint_{\mathbb{R}^4}\mathbb{T}P(\Delta\beta|\gamma,\beta, \hat{x})P(\gamma,\beta)\phi(\hat{x}) d\hat{x} d\Delta\beta d\gamma d\beta \\
 & =   \iint_{\mathbb{R}^2}P(\gamma,\beta)(\Phi(\frac{\beta}{\gamma}) + 0 + \Phi(\frac{\beta}{\gamma})\phi^\prime(\frac{\beta}{\gamma})\frac{1}{2\gamma^2}\eta^2c^2) d\gamma d\beta\\
     & = \mathbb{E}_{\gamma, {\beta}}(\Phi(\frac{{\beta}}{\gamma})) + \frac{\eta^2c^2}{2} \iint_{\mathbb{R}^2}P(\gamma,\beta) \Phi(\frac{\beta}{\gamma})\phi^\prime(\frac{\beta}{\gamma})\frac{1}{\gamma^2} d\gamma d\beta
     \end{aligned}
\end{equation}

The integration of $\mathbb{S}$ is trickier, following the integration of $\mathbb{T}$ and Lemma \ref{scalelemma}

\begin{equation}
    \begin{aligned}
     & \iiiint_{\mathbb{R}^4}\mathbb{S}P(\Delta\beta|\gamma,\beta, \hat{x})P(\gamma,\beta)\phi(\hat{x}) d\hat{x} d\Delta\beta d\gamma d\beta \\
     & = \iiint_{\mathbb{R}^3}\phi(\frac{\beta}{\gamma})\Delta\beta (-\frac{\beta}{\gamma^2})P(\gamma,\beta) \int_{\mathbb{R}} \hat{x}P(\Delta\beta|\gamma,\beta, \hat{x})\phi(\hat{x})d\hat{x}d\Delta\beta d\gamma d\beta\\
     & - \iiint_{\mathbb{R}^3}\Delta\beta^2 (\frac{\phi(\frac{\beta}{\gamma})}{\gamma^2} + \frac{\phi^\prime(\frac{\beta}{\gamma})\beta}{\gamma^3} )P(\gamma,\beta) \int_{\mathbb{R}} \hat{x}P(\Delta\beta|\gamma,\beta, \hat{x})\phi(\hat{x})d\hat{x}d\Delta\beta d\gamma d\beta\\
     & + \iiint_{\mathbb{R}^3}\Delta\beta^2 (\frac{\phi(\frac{\beta}{\gamma})\beta}{\gamma^3} + \frac{\phi^\prime(\frac{\beta}{\gamma})\beta^2}{2\gamma^4} )P(\gamma,\beta) \int_{\mathbb{R}} \hat{x}^2P(\Delta\beta|\gamma,\beta, \hat{x})\phi(\hat{x})d\hat{x}d\Delta\beta d\gamma d\beta \end{aligned}
\end{equation}

For simplicity of notation, let $$g(y) = \int_{-\infty}^{y} \hat{x} \phi(\hat{x})d\hat{x}$$
$$h(y) = \int_{-\infty}^{y} \hat{x}^2 \phi(\hat{x})d\hat{x}$$

Note that $\hat{X}$ follows the standard normal distribution, which has mean $0$ and variance $1$, therefore:

$$\int_{y}^{+\infty} \hat{x} \phi(\hat{x})d\hat{x} = -g(y)$$
$$\int_{y}^{+\infty} \hat{x}^2 \phi(\hat{x})d\hat{x} = 1 - h(y)$$

The integration now becomes

\begin{equation}
\label{IntegrationForS}
    \begin{aligned}
    & \iiiint_{\mathbb{R}^4}\mathbb{S}P(\Delta\beta|\gamma,\beta, \hat{x})P(\gamma,\beta)\phi(\hat{x}) d\hat{x} d\Delta\beta d\gamma d\beta \\
     & = \iiint_{\mathbb{R}^3}\phi(\frac{\beta}{\gamma})\Delta\beta (-\frac{\beta}{\gamma^2})P(\gamma,\beta) [\delta_{\Delta\beta} g(-\frac{\beta}{\gamma}) - \eta^{-1}f(-\frac{\Delta\beta}{\eta}) g(-\frac{\beta}{\gamma})]d\Delta\beta d\gamma d\beta \\
     & - \iiint_{\mathbb{R}^3}\Delta\beta^2 (\frac{\phi(\frac{\beta}{\gamma})}{\gamma^2} + \frac{\phi^\prime(\frac{\beta}{\gamma})\beta}{\gamma^3} )P(\gamma,\beta) [\delta_{\Delta\beta}g(-\frac{\beta}{\gamma}) - \eta^{-1}f(-\frac{\Delta\beta}{\eta})  g(-\frac{\beta}{\gamma})]d\Delta\beta d\gamma d\beta\\
     & + \iiint_{\mathbb{R}^3}\Delta\beta^2 (\frac{\phi(\frac{\beta}{\gamma})\beta}{\gamma^3} + \frac{\phi^\prime(\frac{\beta}{\gamma})\beta^2}{2\gamma^4} )P(\gamma,\beta) [\delta_{\Delta\beta} h(-\frac{\beta}{\gamma}) + \eta^{-1}f(-\frac{\Delta\beta}{\eta})  (1-h(-\frac{\beta}{\gamma}))]d\Delta\beta d\gamma d\beta \\
     & = 0 + \iint_{\mathbb{R}^2} [(\frac{\phi(\frac{\beta}{\gamma})}{\gamma^2} + \frac{\phi^\prime(\frac{\beta}{\gamma})\beta}{\gamma^3} ) g(-\frac{\beta}{\gamma}) + (\frac{\phi(\frac{\beta}{\gamma})\beta}{\gamma^3} + \frac{\phi^\prime(\frac{\beta}{\gamma})\beta^2}{2\gamma^4} ) (1-h(-\frac{\beta}{\gamma}))]P(\gamma,\beta)\eta^2c^2d\gamma d\beta \\
     & = \eta^2c^2\iint_{\mathbb{R}^2} [(\frac{\phi(\frac{\beta}{\gamma})}{\gamma^2} + \frac{\phi^\prime(\frac{\beta}{\gamma})\beta}{\gamma^3} ) g(-\frac{\beta}{\gamma}) + (\frac{\phi(\frac{\beta}{\gamma})\beta}{\gamma^3} + \frac{\phi^\prime(\frac{\beta}{\gamma})\beta^2}{2\gamma^4} ) (1-h(-\frac{\beta}{\gamma}))]P(\gamma,\beta)d\gamma d\beta \\
     & =  \eta^2c^2\iint_{\mathbb{R}^2}\frac{1}{\gamma^2}[(\phi(\frac{\beta}{\gamma}) + \frac{\phi^\prime(\frac{\beta}{\gamma})\beta}{\gamma} ) g(-\frac{\beta}{\gamma}) + (\frac{\phi(\frac{\beta}{\gamma})\beta}{\gamma} + \frac{\phi^\prime(\frac{\beta}{\gamma})\beta^2}{2\gamma^2} ) (1-h(-\frac{\beta}{\gamma}))]P(\gamma,\beta)d\gamma d\beta 
    \end{aligned}
\end{equation}

Note that $g(-y)$ and $1-h(-y)$ can be explicitly calculated:

\begin{equation}
    \begin{aligned}
    g(-y) & = \int_{-\infty}^{-y} \hat{x} \phi(\hat{x})d\hat{x}  \\
          & = \int_{-\infty}^{-y} s \frac{1}{\sqrt{2\pi}} e ^{-\frac{s^2}{2}}ds \\
          & =  -\frac{1}{\sqrt{2\pi}}  e ^{-\frac{y^2}{2}} \\
          & = -\phi(y) \\
    1-h(-y) & = \int_{-y}^{+\infty} s^2 \phi(s)ds  \\
          & = \int_{-y}^{+\infty} s^2 \frac{1}{\sqrt{2\pi}} e ^{-\frac{s^2}{2}}ds \\
          & = \frac{1}{\sqrt{2\pi}}[(-s e^{-\frac{s^2}{2}})|^{+\infty}_{-y} + \int_{-y}^{+\infty} e^{-\frac{s^2}{2}}ds],  \text{by L'Hospital's Rule} \\
          & = -y\phi(y) + 1 - \Phi(-y) \\
          & = -y\phi(y) + \Phi(y)
    \end{aligned}
\end{equation}

Equation (\ref{IntegrationForS}) now becomes:

\begin{equation}
    \begin{aligned}
     & \iiiint_{\mathbb{R}^4}\mathbb{S}P(\Delta\beta|\gamma,\beta, \hat{x})P(\gamma,\beta)\phi(\hat{x}) d\hat{x} d\Delta\beta d\gamma d\beta \\ 
     & =  \frac{\eta^2c^2}{2}\iint_{\mathbb{R}^2}\frac{1}{\gamma^2}[(2\phi(\frac{\beta}{\gamma}) + \frac{2\phi^\prime(\frac{\beta}{\gamma})\beta}{\gamma} ) (-\phi(\frac{\beta}{\gamma})) + (\frac{2\phi(\frac{\beta}{\gamma})\beta}{\gamma} + \frac{\phi^\prime(\frac{\beta}{\gamma})\beta^2}{\gamma^2} ) (-\frac{\beta}{\gamma}\phi(\frac{\beta}{\gamma}) + \Phi(\frac{\beta}{\gamma}))]P(\gamma,\beta)d\gamma d\beta 
    \end{aligned}
\end{equation}

Summing up, we get that

$$
\mathbb{E}_{\gamma^\prime, {\beta}^\prime}(\Phi(\frac{{\beta}^\prime}{\gamma^\prime})) = \mathbb{E}_{\gamma, {\beta}}(\Phi(\frac{{\beta}}{\gamma})) + \frac{\eta^2c^2}{2}\iint_{R^2}\frac{1}{\gamma^2} K(\frac{\beta}{\gamma})P(\gamma,\beta) d\gamma d\beta
$$

where since $\phi^\prime(x) = (-x)\phi(x)$
\begin{equation}
    \begin{aligned}
    K(\frac{\beta}{\gamma}) &= (2\phi(\frac{\beta}{\gamma}) + \frac{2\phi^\prime(\frac{\beta}{\gamma})\beta}{\gamma} ) (-\phi(\frac{\beta}{\gamma})) + (\frac{2\phi(\frac{\beta}{\gamma})\beta}{\gamma} + \frac{\phi^\prime(\frac{\beta}{\gamma})\beta^2}{\gamma^2} ) (-\frac{\beta}{\gamma}\phi(\frac{\beta}{\gamma}) + \Phi(\frac{\beta}{\gamma})) + \Phi(\frac{\beta}{\gamma})\phi^\prime(\frac{\beta}{\gamma}) \\
    & = (2\phi(\frac{\beta}{\gamma}) - 2 \phi(\frac{\beta}{\gamma})(\frac{\beta}{\gamma})^2) (-\phi(\frac{\beta}{\gamma})) + (2\phi(\frac{\beta}{\gamma})\frac{\beta}{\gamma} - \phi(\frac{\beta}{\gamma})(\frac{\beta}{\gamma})^3)(-\frac{\beta}{\gamma}\phi(\frac{\beta}{\gamma})
    + \Phi(\frac{\beta}{\gamma}))
    - \Phi(\frac{\beta}{\gamma})\phi(\frac{\beta}{\gamma}) \frac{\beta}{\gamma} \\
    & = 2((\frac{\beta}{\gamma})^2 - 1) \phi(\frac{\beta}{\gamma})^2 + ((\frac{\beta}{\gamma})^4 - 2 (\frac{\beta}{\gamma})^2)\phi(\frac{\beta}{\gamma})^2 + (\frac{\beta}{\gamma}-(\frac{\beta}{\gamma})^3)\phi(\frac{\beta}{\gamma})\Phi(\frac{\beta}{\gamma})\\
    & =  ((\frac{\beta}{\gamma})^4 - 2) \phi(\frac{\beta}{\gamma})^2 + (\frac{\beta}{\gamma}-(\frac{\beta}{\gamma})^3)\phi(\frac{\beta}{\gamma})\Phi(\frac{\beta}{\gamma})
    \end{aligned} 
\end{equation}

\begin{equation}
    \begin{aligned}
    \iint_{\mathbb{R}^2}\frac{1}{\gamma^2} K(\frac{\beta}{\gamma})P(\gamma,\beta) d\gamma d\beta  
    &= \int_{\mathbb{R}}\frac{1}{\gamma^2} P(\gamma) \int_{\mathbb{R}} K(\frac{\beta}{\gamma})P(\beta) d\beta d\gamma \\
    &= \int_{\mathbb{R}}\frac{1}{\gamma^2} P(\gamma) J(\gamma) d\gamma \\
  J(\gamma) &=  \int_{\mathbb{R}} K(\frac{\beta}{\gamma})P(\beta) d\beta
    \end{aligned}
\end{equation}

\begin{figure}[ht!]
\centering
\includegraphics[width=0.5\linewidth]{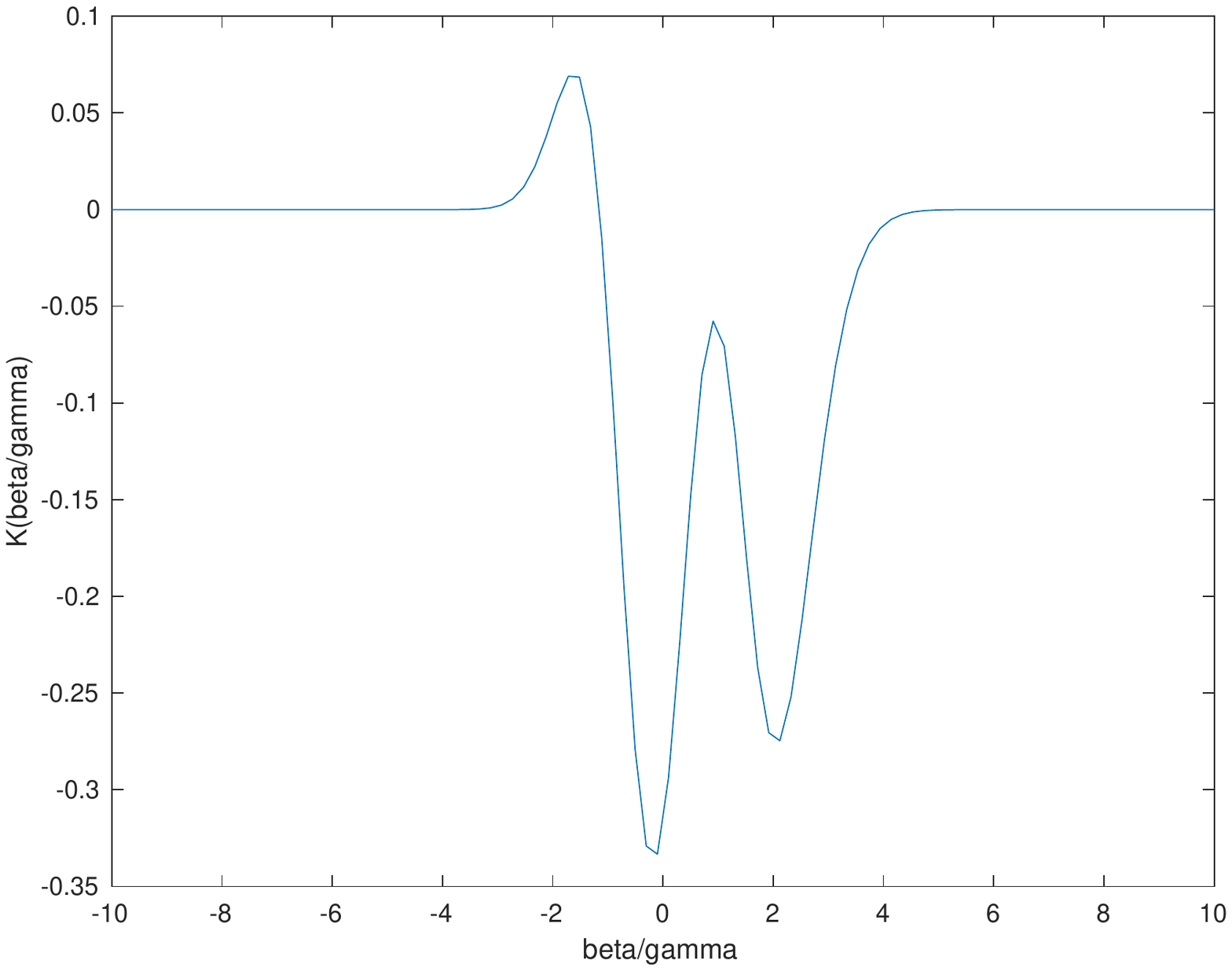}
\caption{The function $K(x)$ \label{K}}
\end{figure}

This value cannot be easily calculated without knowing the distribution function of $\gamma$ and $\beta$, but we can plot the function $K(x)$ in order to closely examine its properties. As shown in Figure \ref{K}, it can be observed that when $K(x)$ is only positive for a very small range of $x < 0$. If we assume that $\beta$ is evenly distributed, for example, when $\beta$ is uniformly distributed on [-1,1], or when $\beta$ is normally distributed, then $J(\gamma) < 0, \forall \gamma$. Note that $\gamma^{-2}P(\gamma) > 0, \forall \gamma$. Hence  $$\iint_{\mathbb{R}^2}\frac{1}{\gamma^2} K(\frac{\beta}{\gamma})P(\gamma,\beta) d\gamma d\beta  < 0 $$ and we prove that 
$$
\mathbb{E}_{\gamma^\prime, {\beta}^\prime}(\Phi(\frac{{\beta}^\prime}{\gamma^\prime})) < \mathbb{E}_{\gamma, {\beta}}(\Phi(\frac{{\beta}}{\gamma}))
$$

\section{Detailed experimental settings}

\subsection{Comparing network with and without BN}

To study the effect of BN on filter collapse, we consider a smaller VGG-like model with 11 layers.
Note that a relatively shallow network is chosen since deeper networks are difficult to train without using batch normalization.
Two variants of the model are trained, one without any batch normalization (VGG11) and one with batch normalization inserted after each convolutional layer and before ReLU activation (VGG11-BN).
Both networks are trained for two rounds and the initial learning rate of each round is reduced to $0.05$ since $0.1$ is too large for VGG11 to converge.
The sparsity cannot be compared by using $\gamma$ due to the absence of BN in VGG11.
Instead, we calculate the $L_1$ norm of each convolutional filter, and a nearly-zero value indicates that the entire filter can be readily removed.

\subsection{Filter collapse as a pruning method}

In the main text, we compare four different pruning methods: multi-round training (\textbf{Retrain}), using large learning rate (\textbf{Increase LR}), using Adam optimizer (\textbf{Adam}) and uniformly shrinking all convolutional layers (\textbf{Uniform filter ablation}). The details of the methods are as follows. For \textbf{Retrain}, the network is trained for multiple rounds using the same strategy. Each round uses the converged model from the previous round as initialization and the network is trained for 9 rounds in total. For \textbf{Increase LR}, we choose initial learning rate in range $[0.1, 1]$. For \textbf{Adam}, we use initial learning rate $0.001$ and weight decay factor $0.0008$. For \textbf{Uniform filter ablation}, we select a ratio $\alpha$ and construct a new network with the same architecture as the original one, while the number of channels in each layer is $\alpha$ times of the number of channels in the original network. This shrinked/ablated network is then trained from scratch. We use $\alpha\in [0.5, 1]$. All methods involves single round of training except \textbf{Retrain}, and no finetuning is performed.